\documentclass[a4paper]{article}

\usepackage[english]{babel}
\usepackage[utf8]{inputenc}
\usepackage{amsmath}
\usepackage{graphicx}
\usepackage{caption}
\usepackage{amsfonts}
\usepackage{verbatim}
\usepackage[colorinlistoftodos]{todonotes}
\usepackage{authblk}
\usepackage{multicol}
\usepackage{xcolor}
\usepackage{wrapfig}
\usepackage{bm}
\usepackage{dsfont}
\usepackage{hyperref}
\usepackage[a4paper,left=18mm,right=18mm, top=3.5cm, bottom=4cm]{geometry}

\newcommand{\FF}{\mathcal{F}}

\newcommand{\bx}{\mathbf{x}}

\title{Reservoirs learn to learn}

\author[1]{Anand Subramoney}
\author[1]{Franz Scherr}
\author[1]{Wolfgang Maass}
\affil[1]{Institute for Theoretical Computer Science, Graz University of Technology, Austria}

\date{\today}

\begin{document}
\maketitle

\begin{abstract}
   The common procedure in reservoir computing is to take a ``found'' reservoir, such as a recurrent neural network with randomly chosen synaptic weights or a complex physical device, and to adapt the weights of linear readouts from this reservoir for a particular computing task. 
   We address the question of whether the performance of reservoir computing can be significantly enhanced if one instead optimizes some (hyper)parameters of the reservoir, not for a single task but for the range of all possible tasks in which one is potentially interested, before the weights of linear readouts are optimized for a particular computing task. 
   After all, networks of neurons in the brain are also known to be not randomly connected. 
   Rather, their structure and parameters emerge from complex evolutionary and developmental processes, arguably in a way that enhances speed and accuracy of subsequent learning of any concrete task that is likely to be essential for the survival of the organism.  
   We apply the Learning-to-Learn (L2L) paradigm to mimick this two-tier process, where a set of (hyper)parameters of the reservoir are optimized for a whole family of learning tasks. 
   We found that this substantially enhances the performance of reservoir computing for the families of tasks that we considered. 
   Furthermore, L2L enables a new form of reservoir learning that tends to enable even faster learning, where not even the weights of readouts need to be adjusted for learning a concrete task. 
   We present demos and performance results of these new forms of reservoir computing for reservoirs that consist of networks of spiking neurons, and are hence of particular interest from the perspective of neuroscience and implementations in spike-based neuromorphic hardware. 
   We leave it as an open question what performance advantage the new methods that we propose provide for other types of reservoirs.
\end{abstract}

\section{Introduction}
One motivation for the introduction of the liquid computing model \cite{maass_real-time_2002} was to understand how complex neural circuits in the brain, or cortical columns, are able to support the diverse computing and learning tasks which the brain has to solve. 
It was shown that recurrent networks of spiking neurons (RSNNs) with randomly chosen weights, including models for cortical columns with given connection probability between laminae and neural populations, could in fact support a large number of different learning tasks, where only the synaptic weights to readout neurons were adapted for a specific task \cite{maass2004fading, haeusler2006statistical}. Independently from that, a similar framework~\cite{jaeger_echo_2001} was developed for artificial neural networks, and both methods were subsumed under the umbrella of reservoir computing~\cite{reservoircomputing}. Our methods for training reservoirs that are discussed in this paper have so far only been tested for reservoirs consisting of spiking neurons, as in the liquid computing model.

Considering the learning capabilities of the brain, it is fair to assume that synaptic weights of these neural networks are not just randomly chosen, but shaped through a host of processes -- from evolution, over development to preceding learning experiences. 
These processes are likely to aim at improving the learning and computing capability of the network. 
Hence we asked whether the performance of reservoirs can also be improved by optimizing the weights of recurrent connections within the recurrent network for a large range of learning tasks. 
The Learning-to-Learn (L2L) setup offers a suitable framework for examining this question. 
This framework builds on a long tradition of investigating L2L, also referred to as meta-learning, in cognitive science, neuroscience, and machine learning \cite{abraham1996metaplasticity,wang_prefrontal_2018,hochreiter2001learning,wang2016learning}.
The formal model from \cite{hochreiter2001learning,wang2016learning} and related recent work in machine learning assumes that learning (or optimization) takes place in two interacting loops (see figure~\ref{fig:ltl_schematic}A).
The outer loop aims at capturing the impact of adaptation on a larger time scale (such as evolution, development, and prior learning in the case of brains). 
It optimizes a set of parameters  $\Theta$, for a -- in general infinitely large -- family $\FF$ of learning tasks. 
Any learning or optimization method can be used for that. 
For learning a particular task $C$ from $\FF$ in the inner loop, the neural network can adapt those of its parameters which do not belong to the hyperparameters $\Theta$ that are controlled by the outer loop. 
These are in our first demo (section~\ref{sec:ltl-readout}) the weights of readout neurons.
In our second demo in section~\ref{sec:ltl-static} we assume that -- like in \cite{wang_prefrontal_2018,wang2016learning,hochreiter2001learning} --  ALL weights from, to, and within the neural network, in particular also the weights of readout neurons, are controlled by the outer loop. 
In this case just the dynamics of the network can be used to maintain information from preceding examples for the current learning task in order to  produce a desirable output for the current network input. 
One exciting feature of this L2L approach is that all synaptic weights of the network can be used to encode a really efficient network learning algorithm. 
It was recently shown in \cite{bellec_etal_nips_2018} that this form of L2L can also be applied to RSNNs. 
We discuss in section~\ref{sec:ltl-static} also the interesting fact that L2L induces priors and internal models into reservoirs.

The structure of this article is as follows. 
We address in section~\ref{sec:ltl-readout} the first form of L2L, where synaptic weights to readout neurons can be trained for each learning task, exactly like in the standard reservoir computing paradigm. 
We discuss in section~\ref{sec:ltl-static} the more extreme form of L2L where ALL synaptic weights are determined by the outer loop of L2L, so that no synaptic plasticity is needed for learning in the inner loop. 
In section~\ref{sec:methods} we give full technical details for the demos given in sections \ref{sec:ltl-readout} and \ref{sec:ltl-static}. 
Finally, in section~\ref{sec:discussion} we will discuss implications of these results, and list a number of related open problems.

\section{Optimizing reservoirs to learn}\label{sec:ltl-readout}

\begin{figure}
	\centering
	\includegraphics[width=\textwidth]{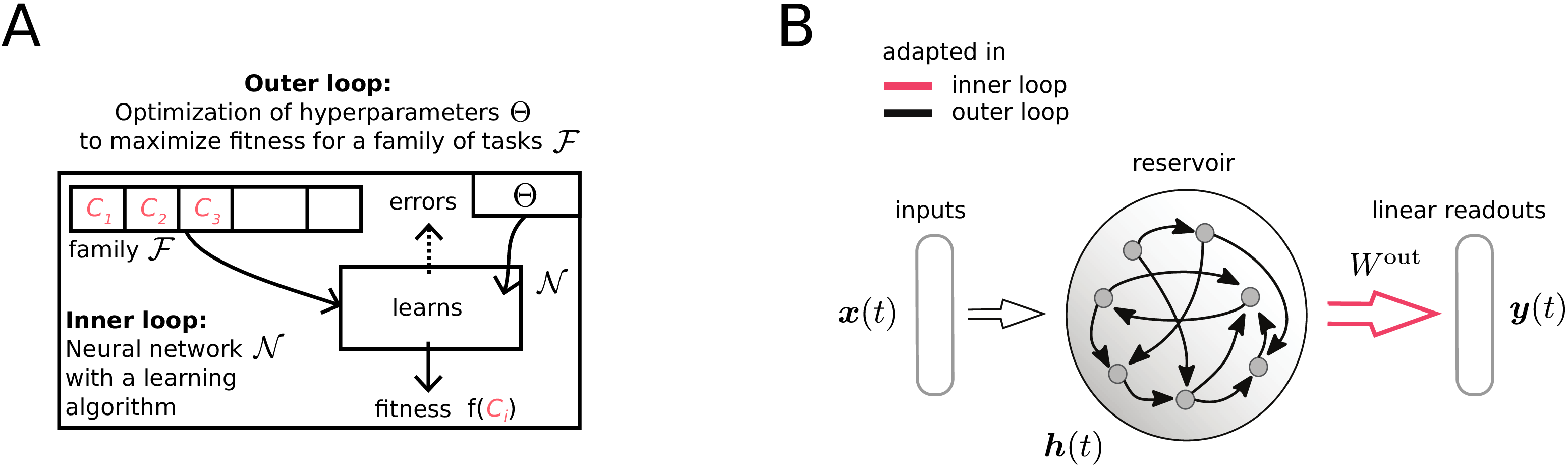}
	\caption{\textbf{Learning-to-Learn setup}: \textbf{A}) Schematic of the nested optimization that is carried out in Learning-to-Learn (L2L). \textbf{B}) Learning architecture that is used to obtain optimized reservoirs}
	\label{fig:ltl_schematic}
\end{figure}

In the typical workflow of solving a task in reservoir computing, we have to address two main issues 1) a suitable reservoir has to be generated and 2) a readout function has to be determined that maps the state of the reservoir to a target output.
In the following, we address the first issue by a close investigation of how we can improve the process of obtaining suitable reservoirs.
For this purpose, we consider here RSNNs as the implementation of the reservoir and its state refers to the activity of all units within the network.
In order to generate an instance of such a reservoir, one usually specifies a particular network architecture of the RSNN and then generates the corresponding synaptic weights at random. Those remain fixed throughout learning of a particular task. 
Clearly, one can tune this random creation process to better suit the needs of the considered task. 
For example, one can adapt the probability distribution from which weights are drawn. 
However, it is likely that a reservoir, generated according to a coarse random procedure, is far from perfect at producing reservoir states that are really useful for the readout. 

A more principled way of generating a suitable reservoir is to optimize their dynamics for the range of tasks to be expected, such that a readout can easily extract the information it needs.

\paragraph{Description of optimized reservoirs:}
The main characteristic of our approach is to view the weight of every synaptic connection of the RSNN that implements the reservoir as hyperparameters $\Theta$, and to optimize them for the range of tasks. In particular, $\Theta$ includes both recurrent and input weights ($W^\mathrm{rec}$, $W^\mathrm{in}$), but also the initialization of the readout $W^{\mathrm{out, init}}$.
This viewpoint allows us to tune the dynamics of the reservoir to give rise to particularly useful reservoir states. 
Learning of a particular task can then be carried out as usual, where commonly a linear readout is learned, for example by the method of least squares or even simpler per gradient descent.

\begin{figure}
	\centering
	\includegraphics[width=\textwidth]{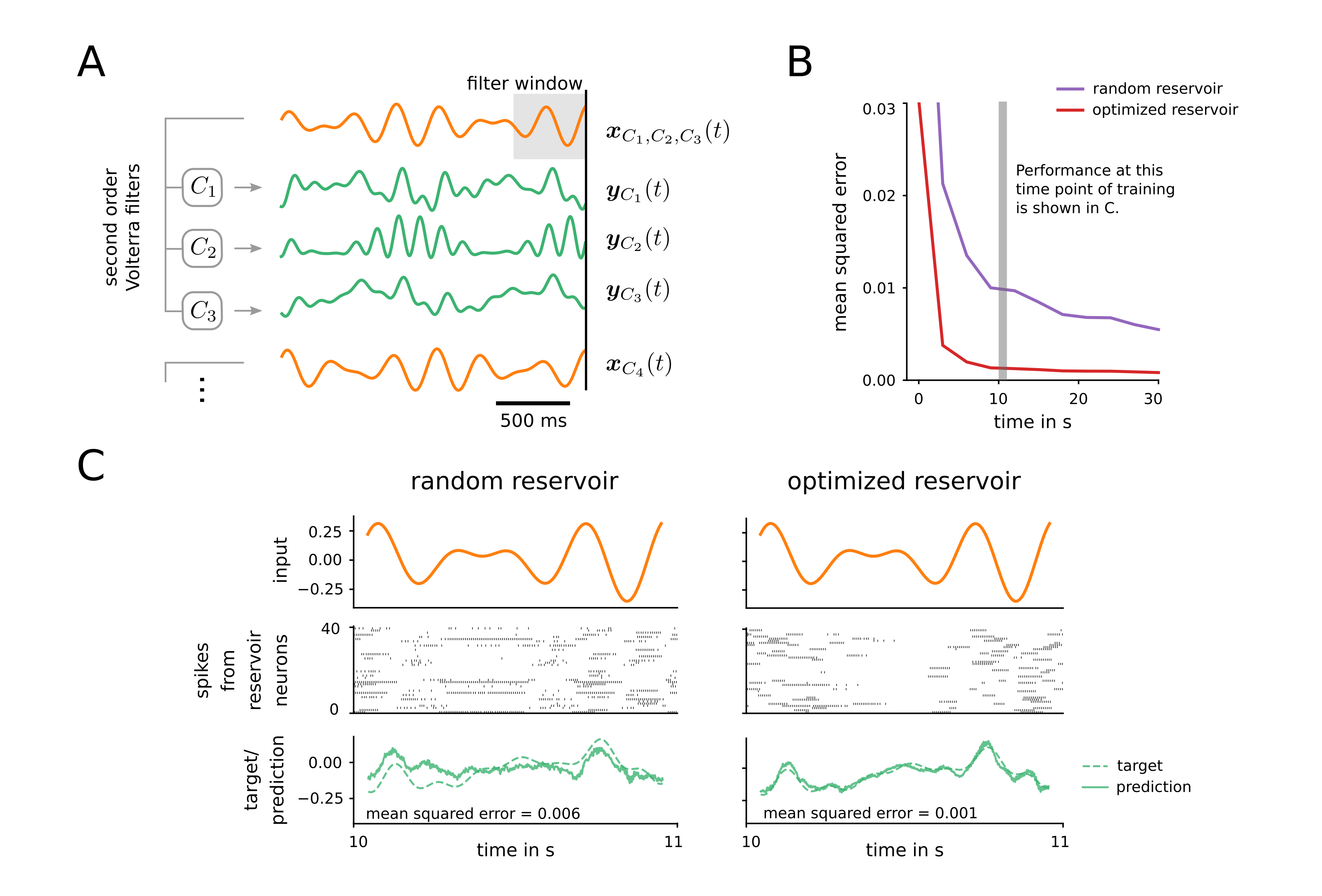}
	\caption{\textbf{Learning to learn a nonlinear transformation of a time series}: \textbf{A}) Different tasks $C_i$ arise by sampling second order Volterra kernels according to a random procedure. Input time series $\bm x_C(t)$ are given as a sum of sines with random properties. To exhibit the variability in the Volterra kernels, we show three examples where different Volterra kernels are applied to the same input. \textbf{B}) Learning performance in the inner loop using the learning rule \eqref{eq:grad_desc}, both for the case of a reservoir with random weights, and for a reservoir that was trained in the outer loop by L2L. Performance at the indicated time window is shown in Panel C. \textbf{C}) Sample performance of a random reservoir and of a optimized reservoir after readouts have been trained for $10$ seconds. Network activity shows 40 neurons out of 800.}
	\label{fig:liquid}
\end{figure}

As previously described, two interacting loops of optimization are introduced, consists of an inner loop and an outer loop (figure~\ref{fig:ltl_schematic}A). The inner loop consists here of tasks $C$ that require to map an input time series $\bm x_C(t)$ to a target time series $\bm y_C(t)$ (see figure~\ref{fig:liquid}A). 
To solve such tasks, $\bm x_C(t)$ is passed as a stream to the reservoir, which then processes these inputs, and produces reservoir states $\bm h_C(t)$. 
The emerging features are then used for target prediction by a linear readout: 
\begin{equation}
	\widehat{\bm y}_C(t) = W_C^\mathrm{out} [\bm x_C(t) ,  \bm h_C(t)]^T~.
\end{equation}
On this level of the inner loop, only the readout weights $W_C^\mathrm{out}$ are learned. 
Specifically, we chose here a particularly simple plasticity rule acting upon these weights, given by gradient descent:
\begin{equation}
	\Delta W^\mathrm{out}_C = \eta \Big(\bm y_C(t) - \widehat{\bm y}_C(t)\Big) \cdot \bm h_C(t)^T~, \label{eq:grad_desc}
\end{equation}
which can be applied continuously, or changes can be accumulated. Note that the initialization of readout weights is provided as a hyperparameter $W^\mathrm{out,init}$ and $\eta$ represents a learning rate. 

On the other hand, the outer loop is concerned with improving the learning process in the inner loop for an entire family of tasks $\mathcal{F}$. This goal is formalized using an optimization objective that acts upon the hyperparameters $\Theta = \{W^\mathrm{in}, W^\mathrm{rec}, W^\mathrm{out,init}\}$:
\begin{eqnarray}
	\min_{\Theta} && \mathds{E}_{C \sim \mathcal{F}} \Bigg[\int_t \Big\|\bm y_C(t) - \widehat{\bm y}_C(t)\Big\|_2^2\Bigg] \label{eq:outer_loop}\\
	\text{subject to} && \Delta W^\mathrm{out}_C = \eta \Big(\bm y_C(t) - \widehat{\bm y}_C(t)\Big) \cdot \bm h_C(t)^T \quad (\text{readout learning})
\end{eqnarray}

\paragraph{Regressing Volterra filters:}
Models of reservoir computing typically get applied to tasks that exhibit nontrivial temporal relationships in the mapping from input signal $\bm x_C(t)$ to target $\bm y_C(t)$. Such tasks are suitable because reservoirs have a property of fading memory: Recent events leave a footprint in the reservoir dynamics which can later be extracted by appropriate readouts. 
Theory guarantees that a large enough reservoir can retain all relevant information. 
In practice, one is bound to a dynamical system of a limited size and hence, it is likely that a reservoir, optimized for the memory requirements and time scales of the specific task family at hand, will perform better than  a reservoir which was generated at random.

We consider a task family $\FF$ where each task $C$ is determined by a randomly chosen Volterra filter~\cite{volterra2005theory}.
Here, the target $\bm y_C(t)$ arises by application of a randomly chosen second order Volterra filter~\cite{volterra2005theory} to the input $\bm x_C(t)$:
\begin{eqnarray}
	\bm y_C(t) = \int_\tau k_C^1(\tau) \bm x_C(t - \tau)\, d\tau + \int_{\tau_1} \int_{\tau_2} k_{C}^2(\tau_1, \tau_2) \bm x_C(t - \tau_1) \bm x_C(t - \tau_2)\, d\tau_1 d\tau_2~, \label{eq:volterra}
\end{eqnarray}
see figure~\ref{fig:liquid}A. The input signal $\bm x_C(t$ is given as a sum of two sines with different frequencies and with random phase and amplitude.
The kernel used in the filter is also sampled randomly according to a predefined procedure for each task $C$, see methods~\ref{sec:liquid_methods}, and exhibits a typical temporal time scale. 
Here, the reservoir is responsible to provide suitable features that typically arise for such second order Volterra filters. In this way readout weights $W_C^\mathrm{out}$, which are adapted according to equation~\eqref{eq:grad_desc}, can easily extract the required information.

\paragraph{Implementation:} \quad
The simulations were carried out in discrete time, with steps of $1\,\mathrm{ms}$ length. 
We used a network of $800$ recurrently connected neurons with leaky integrate-and-fire (LIF) dynamics. Such neurons are equipped with a membrane potential in which they integrate input current. If this potential crosses a certain threshold, they emit a spike and the membrane voltage is reset, see methods~\ref{sec:lif} for details.
The reservoir state was implemented as a concatenation of the exponentially filtered spike trains of all neurons (with a time constant of $\tau_\mathrm{readout} = 20\,\mathrm{ms}$).
Learning of the linear readout weights in the inner loop was implemented using gradient descent as outlined in equation~\eqref{eq:grad_desc}. We accumulated weight changes in chunks of $1000\,\mathrm{ms}$ and applied them at the end.
The objective for the outer loop, as given in equation~\eqref{eq:outer_loop}, was optimized using backpropagation through time (BPTT), which is an algorithm to perform gradient descent in recurrent neural networks. 
Observe that this is possible because the dynamics of the  plasticity in equation~\eqref{eq:grad_desc} is itself differentiable, and can therefore be optimized by gradient descent. 
Because the threshold function that determines the neuron outputs is not differentiable, a heuristic was required to address this problem. Details can be found in the methods~\ref{sec:bptt}.

\paragraph{Results:} \quad
The reservoir that emerged from outer-loop training was compared against a reference baseline, whose weights were not optimized for the task family, but had otherwise exactly the same structure and learning rule for the readout.
In figure~\ref{fig:liquid}B we report the learning performance on unseen task instances from the family $\mathcal{F}$, averaged over 200 different tasks. 
We find that the learning performance of the optimized reservoir is substantially improved as compared to the random baseline. 

This becomes even more obvious when one compares the quality of the fit on a concrete example as shown in figure~\ref{fig:liquid}C. 
Whereas the random reservoir fails to make consistent predictions about the desired output signal based on the reservoir state, the optimized reservoir is able to capture all important aspects of the target signal. 
This occurred occurs just $10$ seconds within learning the specific task, because the optimized reservoir was already confronted before with tasks of a similar structure, and could capture through the outer loop optimization the smoothness of the Volterra kernels and the relevant time dependencies in its recurrent weights.

\section{Reservoirs can also learn without changing synaptic weights to readout neurons}\label{sec:ltl-static}

\begin{figure}
	\begin{center}
        \includegraphics[width=\textwidth]{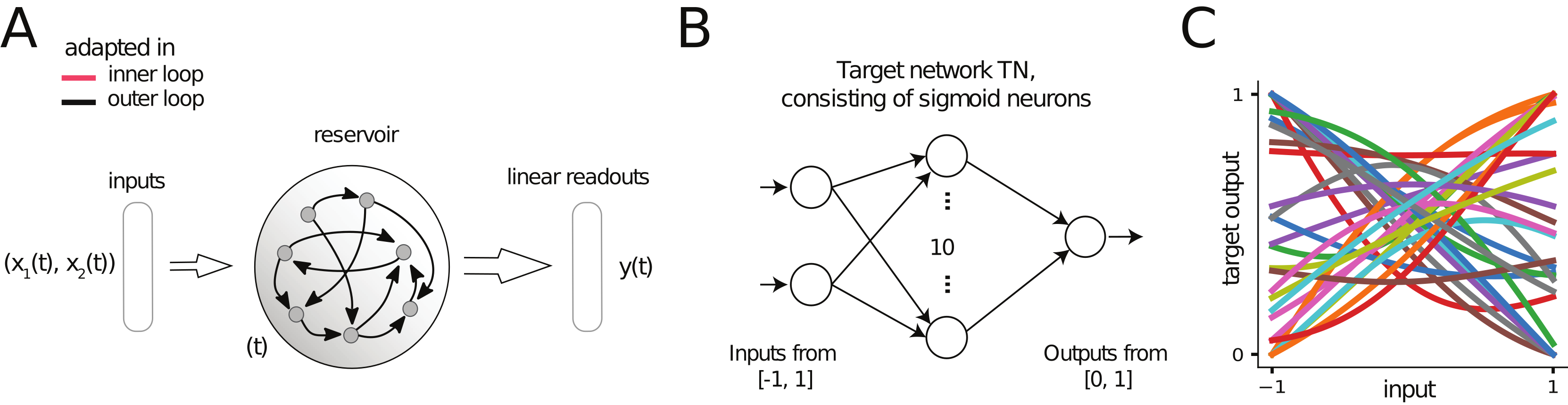}
        \caption{\label{fig:ltl2-setup} \textbf{L2L setup with reservoirs that learn using their internal dynamics}
            \textbf{A}) Learning architecture for RSNN reservoirs.  All the weights are only updated in the outer-loop training using BPTT.
            \textbf{B}) Supervised regression tasks are implemented as neural networks with randomly sampled weights: target networks (TN).
            \textbf{C}) Sample input/output curves of TNs on a 1D subset of the 2D input space, for different weight and bias values.
        }
	\end{center}
\end{figure}

\begin{figure}
	\begin{center}
        \includegraphics[width=\textwidth]{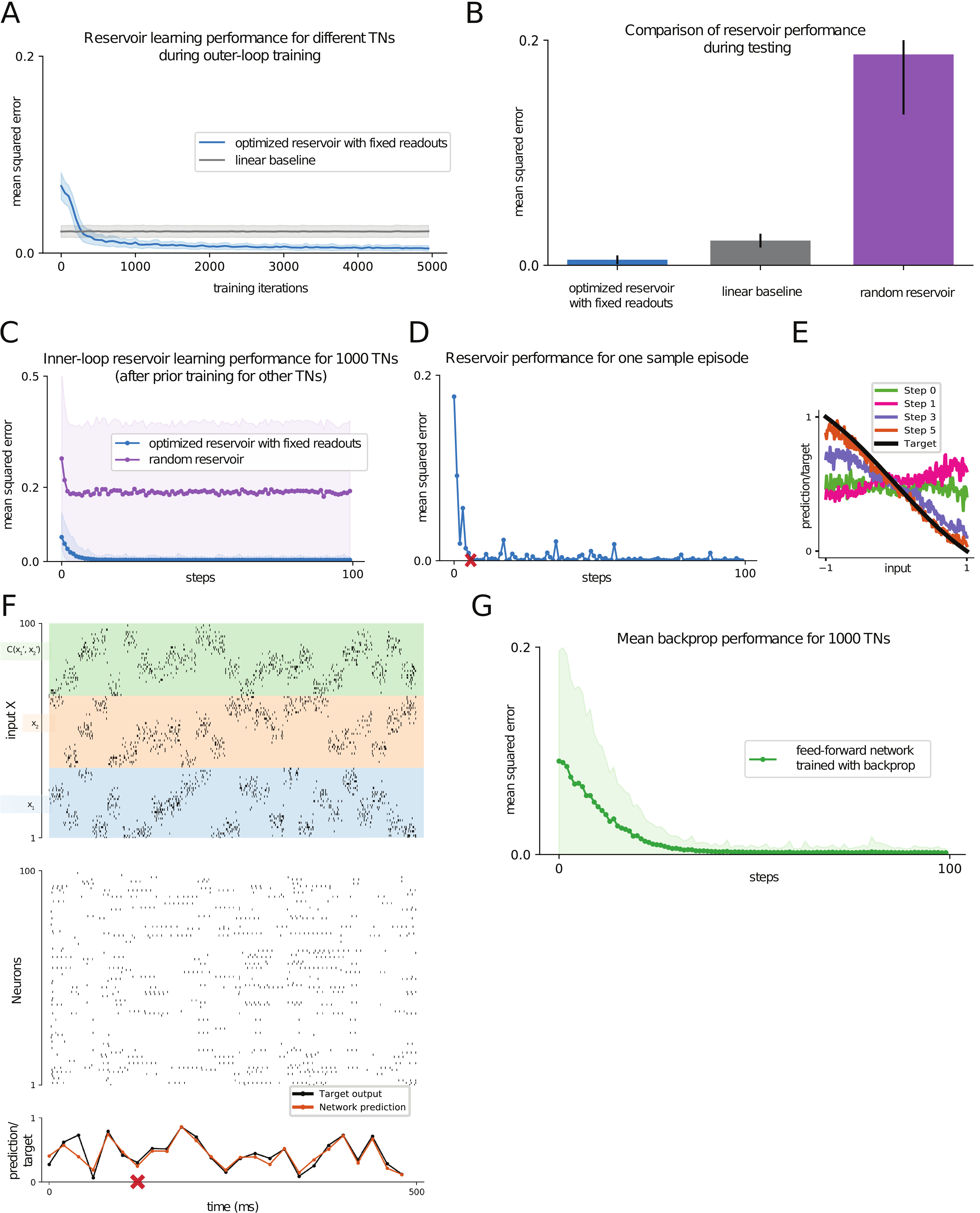}
	  \caption{(Caption next page.)}
	\end{center}
\end{figure}
\addtocounter{figure}{-1}
\begin{figure}
    \caption{\label{fig:ltl2-results} \textbf{Learning to learn a nonlinear function that is defined by an unknown target network (TN):}
	\textbf{A}) Performance of the reservoir in learning a new TN during training in the outer loop of L2L.
	\textbf{B}) Performance of the optimized reservoir during testing compared to a random reservoir and the linear baseline.
	\textbf{C}) Learning performance within a single inner-loop episode of the reservoir for 1000 new TNs (mean and one standard deviation). Performance is compared to that of a random reservoir.
	\textbf{D}) Performance for a single sample TN, a red cross marks the step after which output predictions became very good for this TN. The spike raster for this learning process is the one depicted in (F). 
	\textbf{E}) The internal model of the reservoir (as described in the text) is shown for the first few steps of inner loop learning. The reservoir starts by predicting a smooth function, and updates its internal model in just $5$ steps to correctly predict the target function.
    \textbf{F}) Network input (top row, only 100 of 300 neurons shown), internal spike-based processing with low firing rates in the neuron populations (middle row), and network output (bottom row) for 25 steps of 20 ms each. 
	\textbf{G}) Learning performance of backpropagation for the same 1000 TNs as in C, working directly on the ANN from figure~\ref{fig:ltl2-setup}B, with a prior for small weights, with the best hyper-parameters from a grid-search.
        }
\end{figure}
We next asked whether reservoirs could also learn a specific task without changing any synaptic weight, not even weights to readout neurons.
It was shown in~\cite{hochreiter2001learning} that LSTM networks can learn nonlinear functions from a teacher without modifying their recurrent or readout weights.
It has recently been argued in~\cite{wang_prefrontal_2018} that the pre-frontal cortex (PFC) accumulates knowledge during fast reward-based learning in its short-term memory, without using synaptic plasticity, see the text to supplementary figure~3 in \cite{wang_prefrontal_2018}. 
The experimental results of \cite{perich2018neural} also suggest a prominent role of network dynamics and short-term memory for fast learning in the motor cortex.
Inspired by these results from biology and machine learning, we explored the extent to which recurrent networks of spiking neurons can learn using just their internal dynamics, without synaptic plasticity.

In this section, we show that one can generate reservoirs through L2L that are able to learn with fixed weights, provided that the reservoir receives feedback about the prediction target as input.
In addition, relying on the internal dynamics of the reservoir to learn allows the reservoir to learn as fast as possible for a given task i.e. the learning speed is not determined by any predetermined learning rate.

\paragraph{Target networks as the task family $\FF$:}\quad
We chose the task family to demonstrate that reservoirs can use their internal dynamics to regress complex non-linear functions, and are not limited to generating or predicting temporal patterns.
This task family also allows us to illustrate and analyse the learning process in the inner loop more explicitly.
We defined the family of tasks $\FF$ using a family of non-linear functions that are each defined by a target feed-forward network (TN) as illustrated in figure~\ref{fig:ltl2-setup}B.
Specifically, we chose a class of continuous functions of two real-valued variables $(x_1,x_2)$ as the family $\mathcal{F}$ of tasks. 
This class was defined as the family of all functions that can be computed by a 2-layer artificial neural network of sigmoidal neurons with 10 neurons in the hidden layer, and weights and biases in the range [-1, 1].
Thus overall, each such target network (TN) from $\mathcal{F}$ was defined through 40 parameters in the range [-1, 1]: 30 weights and 10 biases. 
Random instances of target networks were generated for each episode by randomly sampling the 40 parameters in the above range.
Most of the functions that are computed by TNs from the class $\mathcal{F}$ are nonlinear, as illustrated in  figure~\ref{fig:ltl2-setup}C for the case of inputs $(x_1, x_2)$ with $x_1  =  x_2$.

\paragraph{Learning setup:}
In an inner loop learning episode, the reservoir was shown a sequence of pairs of inputs ($x_1, x_2$) and delayed targets $C(x_1', x_2')$ sampled from the non-linear function generated by one random instance of the TN.
After each such pair was presented, the reservoir was trained to produce a prediction $\hat{C}(x_1, x_2)$ of $C(x_1, x_2)$.
The task of the reservoir was to produce predictions with a low error.
In other words, the task of the reservoir was to perform non-linear regression on the presented pairs of inputs and targets and produce predictions of low-error on new inputs.
The reservoir was optimized in the outer loop to learn this fast and well.

When giving an input $x_1, x_2$ for which the reservoir had to produce prediction $\hat{C}(x_1, x_2)$, we could not also give the target $C(x_1, x_2)$ for that same input at the same time.
This is because, the reservoir could then ``cheat'' by simply producing this value $C(x_1, x_2)$ as its prediction $\hat{C}(x_1, x_2)$.
Therefore, we gave the target value to the reservoir with a delay, after it had generated the prediction $\hat{C}(x_1, x_2)$.
Giving the target value as input to the reservoir is necessary, as otherwise, the reservoir has no way of figuring out the specific underlying non-linear function for which it needs to make predictions.

Learning is carried out simultaneously in two loops as before (see figure~\ref{fig:ltl_schematic}A).
Like in \cite{hochreiter2001learning,wang2016learning,duan2016rl} we let all synaptic weights of $\mathcal{N}$, including the recurrent, input and readout weights, to belong to the set of hyper-parameters that are optimized in the outer loop.
Hence the network is forced to encode all results from learning the current task $C$ in its internal state, in particular in its firing activity.
Thus the synaptic weights of the neural network $\mathcal{N}$ are free to encode an efficient \textit{algorithm} for learning arbitrary tasks $C$ from $\mathcal{F}$.

\paragraph{Implementation:}
We considered a reservoir $\mathcal{N}$ consisting of $300$ LIF neurons with full connectivity. 
The neuron model is described in the Methods section~\ref{sec:lif}.
All neurons in the reservoir received input from a population $X$ of 300 external input neurons.
A linear readout receiving inputs from all neurons in the reservoir was used for the output predictions.
The reservoir received a stream of 3 types of external inputs (see top row of figure~\ref{fig:ltl2-results}F): the values of $x_1, x_2$, and of the output $C(x_1', x_2')$ of the TN for the preceding input pair $x_1', x_2'$ (set to 0 at the first trial), each represented through population coding in an external population of 100 spiking neurons.
It produced outputs in the form of weighted spike counts during $20$ ms windows from all neurons in the network (see bottom row of figure~\ref{fig:ltl2-results}F). 
The weights for this linear readout were trained, like all weights inside the reservoir, in the outer loop, and remained fixed during learning of a particular TN. 

The training procedure in the outer loop of L2L was as follows: Network training was divided into training episodes.
At the start of each training episode, a new TN was randomly chosen and used to generate target values $C(x_1, x_2) \in [0, 1]$ for randomly chosen input pairs $(x_1,x_2)$.
400 of these input pairs and targets were used as training data, and presented one per step to the reservoir during the episode, where each step lasted $20$ ms.
The reservoir parameters were updated using BPTT to minimize the mean squared error between the reservoir output and the target in the training set, using gradients computed over batches of $10$ such episodes, which formed one iteration of the outer loop.
In other words, each weight update included gradients calculated on the input/target pairs from $10$ different TNs. 
This training procedure forced the reservoir to adapt its parameters in a way that supported learning of many different TNs, rather than specializing on predicting the output of single TN.
After training, the weights of the reservoir remained fixed, and it was required to learn the input/output behaviour of TNs from $\mathcal{F}$ that it had never seen before in an online manner by just using its fading memory and dynamics.
See the Methods (section~\ref{sec:methods-ltl2}) for further details of the implementation.

\paragraph{Results:}
The reservoir achieves low mean-squared error (MSE) for learning new TNs from the family $\mathcal{F}$, significantly surpassing the performance of an optimal linear approximator (linear regression) that was trained on all 400 pairs of inputs and target outputs, see grey bar in figure~\ref{fig:ltl2-results}B. 
One sample of a generic learning process is shown in figure~\ref{fig:ltl2-results}D.

Each sequence of examples evokes an ``internal model'' of the current target function in the internal dynamics of the reservoir.
We make the current internal model of the reservoir visible by probing its prediction $C(x_1,x_2)$ for hypothetical new inputs for evenly spaced points $(x_1, x_2)$ in the entire domain, without allowing it to modify its internal state (otherwise, inputs usually advance the network state according to the dynamics of the network).
Figure~\ref{fig:ltl2-results}E shows the fast evolution of internal models of the reservoir for the TN during the first trials (visualized for a 1D subset of the 2D input space).
One sees that the internal model of the reservoir is from the beginning a smooth function, of the same type as the ones defined by the TNs in $\mathcal{F}$.
Within a few trials this smooth function approximated the TN quite well.
Hence the reservoir had acquired during the training in the outer loop of L2L a prior for the types of functions that are to be learnt, that was encoded in its synaptic weights.
This prior was in fact quite efficient, as Figs.~\ref{fig:ltl2-results}C,D,E show, compared to that of a random reservoir.
The reservoir was able to learn a TN with substantially fewer trials than a generic learning algorithm for learning the TN directly in an artificial neural network as shown in figure~\ref{fig:ltl2-results}G: backpropagation with a prior that favored small weights and biases.
In this case, the target input was given as feedback to the reservoir throughout the episode, and we compare the training error achieved by the reservoir with that of a FF network trained using backpropagation.
A reservoir with a long short-term memory mechanism where we could freeze the memory after low error was achieved allowed us to stop giving the target input after the memory was frozen (results not shown).
This long short-term memory mechanism was in the form of neurons with adapting thresholds as described in~\cite{bellec_etal_nips_2018,bellec2018learncompute}.
These results suggest that L2L is able to install some form of prior knowledge about the task in the reservoir.
We conjectured that the reservoirs fits internal models for smooth functions to the examples it received.

We tested this conjecture in a second, much simpler, L2L scenario.
Here the family $\mathcal{F}$ consisted of all sine functions with arbitrary phase and amplitudes between 0.1 and 5. 
The reservoir also acquired an internal model for sine functions in this setup from training in the outer loop, as shown in~\cite{bellec_etal_nips_2018}.  
Even when we selected examples in an adversarial manner, which happened to be in a straight line, this did not disturb the prior knowledge of the reservoir.

Altogether the network learning that was induced through L2L in the reservoir is of particular interest from the perspective of the design of learning algorithms, since we are not aware of previously documented methods for installing structural priors for online learning of a RSNN.
\section{Methods}\label{sec:methods}

\subsection{Leaky integrate and fire neurons}\label{sec:lif}

We used leaky integrate-and-fire (LIF) models of spiking neurons, where the membrane potential $V_j(t)$ of neuron $j$ evolves according to:
\begin{equation}
    V_j(t+1) = \rho_j\,V_j(t)+(1-\rho_j)\,R_m\,I_j(t) - B_j(t)\,z_j(t)
\end{equation}
where $R_m$ is the membrane resistance, and $\rho_j$ is the decay constant defined using the membrane time constant $\tau_j$ as $\rho_j = e^\frac{-\Delta t }{\tau_{j}}$, where $\Delta t$ is the time step of simulation.
A neuron $j$ spikes as soon at its normalized membrane potential $v_j(t)=\frac{V_j(t)-B_j(t)}{B_j(t)}$ is above its firing threshold $v_\mathrm{th}$.
At each spike time $t$, the membrane potential $V_j(t)$ is reset by subtracting the current threshold value $B_{j}(t)$.
After each spike, the neuron enters a strict refractory period during which it cannot spike.

\subsection{Backpropagation through time}\label{sec:bptt}
We introduced a version of backpropagation through time (BPTT) in \cite{bellec_etal_nips_2018} which allows us to back-propagate the gradient through the discontinuous firing event of spiking neurons.
The firing is formalized through a binary step function $H$ applied to the scaled membrane voltage $v(t)$. 
The gradient is propagated through this step function with a pseudo-derivative as in \cite{courbariaux_binarized_2016,esser_convolutional_2016}, but with a dampened amplitude at each spike.

Specifically, the derivative of the spiking $z_j(t)$ w.r.t to the normalized membrane potential $v_j(t)=\frac{V_j(t)-B_j(t)}{B_j(t)}$ is defined as:
\begin{equation}
    \frac{dz_j(t)}{dv_j(t)} := \gamma\,\text{max}\{0,1-|v_j(t)|\}.\label{eq:bptt}
\end{equation}
In this way the architecture and parameters of a RSNN can be optimized for a given computational task.

\subsection{Optimizing reservoirs to learn} \label{sec:liquid_methods}
\paragraph{Reservoir model:} \quad 
Our reservoir consists of 800 recurrently connected leaky integrate-and-fire (LIF) neurons according to the dynamics defined above. 
The network simulation is carried out in discrete timesteps of $\Delta t = 1\,\mathrm{ms}$. 
The membrane voltage decay was uniform across all neurons and was computed to correspond to a time constant of $20\,\mathrm{ms}$ ($\rho_j = 0.368$). 
The normalized spike threshold was set to $0.02$ and a refractory period of $5\,\mathrm{ms}$ was introduced. 
Synapses had delays of $5\,\mathrm{ms}$.
In the beginning of the experiment, input $W^\mathrm{in}$ and recurrent weights $W^\mathrm{rec}$ were initialized according to Gaussian distributions with zero mean and standard deviations of $\frac{1}{\sqrt{3}}$ and $\frac{1}{\sqrt{800}}$ respectively. 
Similarly, the initial values of the readout $W^\mathrm{out,init}$ were also optimized in the outer loop, and were randomly initialized in the beginning of the experiment according to a uniform distribution, as proposed in~\cite{GlorotBengio:10}.

\paragraph{Readout learning:} \quad
The readout was iteratively adapted according to equation~\eqref{eq:grad_desc}. It received as input the input $\bm x_C(t)$ itself and the features $\bm h_C(t)$ from the reservoir, which were given as exponentially filtered spike trains: $h_{C,j}(t) = \sum_{t' \leq t} \kappa^{t - t'} z_{C,j}(t')$. Here, $\kappa = e^{\frac{-\Delta t}{\tau_\mathrm{readout}}}$ is the decay of leaky readout neurons.
Weight changes were computed at each timestep and accumulated. 
After every second these changes were used to actually modify the readout weights. 
Thus, formulated in discrete time, the plasticity of the readout weights in a task $C$ took the following form:
\begin{equation}
	\Delta W^\mathrm{out}_C = \eta \sum_{t' = t - 1000\,\mathrm{ms}}^{t}\Big(\bm y_C(t') - \widehat{\bm y}_C(t')\Big) \cdot \bm h_C(t')^T, \label{eq:plasticity_readout}
\end{equation}
where $\eta$ is a learning rate.

\paragraph{Outer loop optimization:} \quad
To optimize input and recurrent weights of the reservoir in the outer loop, we simulated the learning procedure described above for $m=40$ different tasks in parallel. 
After each $3$ seconds, the simulation was paused and the outer loop objective was evaluated. Note that the readout weights were updated 3 times within these 3 seconds according to our scheme. The outer loop objective, as given in equation~\ref{eq:outer_loop}, is approximated by:
\begin{eqnarray}
	\mathcal{L} = \frac{1}{m} \sum_{n = 1}^{m} \sum_{t' = t - 2000\,\mathrm{ms}}^{t} \Big\| \bm y_n(t') - \widehat{\bm y}_n(t') \Big\|_2^2 + \mathcal{L}_\mathrm{reg}~.
\end{eqnarray}
We found that learning is improved if one includes only the last two seconds of simulation. This is because the readout weights seem fixed and unaffected by the plasticity of equation~\ref{eq:grad_desc} in the first second, as BPTT cannot see beyond the truncation of 3 seconds. 
The cost function $\mathcal{L}$ was then minimized using a variant of gradient descent (Adam~\cite{kingma2014adam}), where a learning rate of $0.001$ was used. 
The required gradient $\nabla \mathcal{L}$ was computed with BPTT using the $3$ second chunks of simulation and was clipped if the $\mathcal{L}_2$-norm exceeded a value of $1000$.

\paragraph{Regularization:} \quad
In order to encourage the model to settle into a regime of plausible firing rates, we add to the outer loop cost function  a term that penalizes excessive firing rates:
\begin{eqnarray}
	\mathcal{L}_\mathrm{reg} = \alpha \sum_{j = 1}^{800} (f_j - 20\,\mathrm{Hz})^2,\label{eq:regularization}
\end{eqnarray}
with the hyperparameter $\alpha = 1200$. We compute the firing rate of a neuron $f_j$ based on the number of spikes in the past $3$ seconds.

\paragraph{Task details:} \quad
We describe here the procedure according to which the input time series $\bm x_C(t)$ and target time series $\bm y_C(t)$ were generated. 
The input signal was composed of a sum of two sines with random phase $\phi_n \in [0, \frac{\pi}{2}]$ and amplitude $A_n \in [0.5, 1]$, both sampled uniformly in the given interval.
\begin{eqnarray}
	\bm x_C(t) = \sum_{n = 1}^2 A_n \sin(2 \pi \frac{t}{T_n} + \phi_n),
\end{eqnarray}
with periods of $T_1 = 0.323\,\mathrm{s}$ and $T_2 = 0.5\,\mathrm{s}$. 

The corresponding target function $\bm y_C(t)$ was then computed by an application of a random second order Volterra filter to $\bm x_C(t)$ according to equation~\eqref{eq:volterra}. 
Each task uses a different kernel in the Volterra filter and we explain here the process by which we generate the kernels $k^1$ and $k^2$. 
Recall that we truncate the kernels after a time lag of $500\,\mathrm{ms}$. 
Together with the fact that we simulate in discrete time steps of $1\,\mathrm{ms}$ we can represent $k^1$ as a vector with $500$ entries, and $k^2$ as a matrix of dimension $500 \times 500$.

\textit{Sampling $k^1$:} We parametrize $k^1$ as a normalized sum of two different exponential filters with random properties:
\begin{eqnarray}
	\tilde k^1(t) &=& \sum_{n =1}^2 a_n \exp\left(-\frac{t}{b_n}\right)\\
	k^1(t) &=& \frac{\tilde k^1(t)}{\| \tilde k^1 \|_1},
\end{eqnarray}
with $a_n$ being sampled uniformly in $[-1, 1]$, and $b_n$ drawn randomly in $[0.1\,\mathrm{s}, 0.3\,\mathrm{s}]$. For normalization, we use the sum of all entries of the filter in the discrete representation ($t \in \{0, 0.001, 0.002, \dots, 0.499\}$).

\textit{Sampling $k^2$:} We construct $k^2$ to resemble a Gaussian bell shape centered at $t = 0$, with a randomized ``covariance'' matrix $\Sigma$, which we parametrize such that we always obtain a positive definite matrix:
\begin{eqnarray}
	\Sigma = \begin{bmatrix}
		\sqrt{1 + u^2 + v^2} + u & v\\
		v & \sqrt{1 + u^2 + v^2} - u
	\end{bmatrix},
\end{eqnarray}
where $u, v$ are sampled uniformly in $[-12, 12]$. With this we defined the kernel $k^2$ according to:
\begin{eqnarray}
	\tilde k^2(t_1, t_2) &=& \exp\left( - \frac{1}{24} \begin{bmatrix}
		t_1, t_2
	\end{bmatrix}
	\Sigma^{-1}
	\begin{bmatrix}
		t_1 \\
		t_2
	\end{bmatrix}
	 \right)\\
	k^2(t_1, t_2) &=& \frac{\tilde k^2(t_1, t_2)}{ \| \tilde k^2 \|_1} \cdot 14.
\end{eqnarray}
The normalization term here is again given by the sum of all entries of the matrix in the discrete time representation ($[t_1, t_2] \in \{0, 0.001, 0.002, \dots, 0.499\}^2$).

\subsection{Reservoirs can also learn without changing synaptic weights to readout neurons}\label{sec:methods-ltl2}

\paragraph{Reservoir model:} \quad 
The reservoir model used here was the same as that in section~\ref{sec:liquid_methods}, but with $300$ neurons.

\paragraph{Input encoding:} \quad
Analog values were transformed into spiking trains to serve as inputs to the reservoir as follows: 
For each input component, $100$ input neurons are assigned values $m_1,\dots m_{100}$ evenly distributed between the minimum and maximum possible value of the input.
Each input neuron has a Gaussian response field with a particular mean and standard deviation, where the means are uniformly distributed between the minimum and maximum values to be encoded, and with a constant standard deviation.
More precisely, the firing rate $r_i$ (in Hz) of each input neuron $i$ is given by $r_i = r_\mathrm{max}\,\exp\left(-\frac{(m_i - z_i)^2}{2\,\sigma^2}\right)$, where $r_\mathrm{max} = 200$ Hz, $m_i$ is the value assigned to that neuron, $z_i$ is the analog value to be encoded, and $\sigma = \frac{(m_\mathrm{max} - m_\mathrm{min})}{1000}$, $m_\mathrm{min}$ with $m_\mathrm{max}$ being the minimum and maximum values to be encoded.

\paragraph{Setup and training schedule:} \quad
The output of the reservoir was a linear readout that received as input the mean firing rate of each of the neurons per step i.e the number of spikes divided by $20$ for the $20$ ms time window that constitutes a step.

The network training proceeded as follows: A new target function was randomly chosen for each \textit{episode} of training, i.e., the parameters of the target function are chosen uniformly randomly from within the ranges above.

Each \textit{episode} consisted of a sequence of $400$ \textit{steps}, each lasting for $20$ ms.
In each step, one training example from the current function to be learned was presented to the reservoir.
In such a step, the inputs to the reservoir consisted of a randomly chosen vector $\bx = (x_1, x_2)$ as described earlier.
In addition, at each step, the reservoir also got the target value $C(x_1', x_2')$ from the previous step, i.e., the value of the target calculated using the target function for the inputs given at the previous step (in the first step, $C(x_1', x_2')$ is set to $0$).  
The previous target input was provided to the reservoir during all steps of the episode.

All the weights of the reservoir were updated using our variant of BPTT, once per \textit{iteration}, where an \textit{iteration} consists of a batch of $10$ \textit{episodes}, and the weight updates were accumulated across episodes in an iteration.
The ADAM~\cite{kingma2014adam} variant of gradient descent was used with standard parameters and a learning rate of $0.001$.  
The loss function for training was the mean squared error (MSE) of the predictions over an iteration (i.e. over all the steps in an episode, and over the entire batch of episodes in an iteration), with the optimization problem written as:
\begin{equation}
\min_{\Theta} \; \mathds{E}_{C \sim \mathcal{F}} \Bigg[\sum_t \Big(C(x_1^t, x_2^t; \Theta) - \widehat{C}(x_1^t, x_2^t; \Theta)\Big)^2\Bigg]
\end{equation}
In addition, a regularization term was used to maintain a firing rate of $20$ Hz as in equation~\ref{eq:regularization}, with $\alpha = 30$.
In this way, we induce the reservoir to use sparse firing.
We trained the reservoir for $5000$ iterations.

\paragraph{Parameter values:} \quad
The parameters of the leaky integrate-and-fire neurons were as follows: $5$ ms neuronal refractory period, delays spread uniformly between $0-5$ ms, membrane time constant $\tau_j=\tau=20$ms ($\rho_j = \rho = 0.368$) for all neurons $j$, $v_\mathrm{th} = 0.03$ mV baseline threshold voltage. 
The dampening factor for training was $\gamma=0.4$ in equation~\ref{eq:bptt}.

\paragraph{Comparison with Linear baseline:} \quad
The linear baseline was calculated using linear regression with L2 regularization with a regularization factor of $100$ (determined using grid search), using the mean spiking trace of all the neurons.
The mean spiking trace was calculated as follows: First the neuron traces were calculated using an exponential kernel with $20$ ms width and a time constant of $20$ ms.
Then, for every step, the mean value of this trace was calculated to obtain the mean spiking trace.
In figure~\ref{fig:ltl2-results}B, for each episode consisting of $400$ steps, the mean spiking trace from a subset of $320$ steps was used to train the linear regressor, and the mean spiking trace from remaining $80$ steps was used to calculate the test error. The reported baseline is the mean of the test error over one batch of $1000$ episodes with error bars of one standard deviation.

The total test MSE was $0.0056 \pm 0.0039$ (linear baseline MSE was $0.0217 \pm 0.0046$) for the TN task.

\paragraph{Comparison with random reservoir} \quad
In figure~\ref{fig:ltl2-results}B,C, a reservoir with randomly initialized input, recurrent and readout weights was tested in the same way as the optimized reservoir -- with the same sets of inputs, and without any synaptic plasticity in the inner loop.
The plotted curves are the average over 8000 different TNs.

\paragraph{Comparison with backprop:} \quad
The comparison was done for the case where the reservoir was trained on the function family defined by target networks.
A feed-forward (FF) network with $10$ hidden neurons and $1$ output was constructed.
The input to this FF network were the analog values that were used to generate the spiking input and targets for the reservoir.
Therefore the FF had $2$ inputs, one for each of $x_1$ and $x_2$.
The error reported in figure~\ref{fig:ltl2-results}G is the mean training error over $1000$ TNs with error bars of one standard deviation.

The FF network was initialized with Xavier normal initialization \cite{GlorotBengio:10} (which had the best performance, compared to Xavier uniform and plain uniform between $[-1, 1]$).
Adam \cite{kingma2014adam} with AMSGrad \cite{reddi2018convergence} was used with parameters $\eta = 10^{-1}, \beta_1 = 0.7, \beta_2 = 0.9, C = 10^{-5}$. These were the optimal parameters as determined by a grid search.
Together with the Xavier normal initialization and the weight regularization parameter $C$, the training of the FF favoured small weights and biases.

\section{Discussion}\label{sec:discussion}
We have presented a new form of reservoir computing, where the reservoir is optimized for subsequent fast learning of any particular task from a large -- in general even infinitely large -- family of possibly tasks. 
We adapted for that purpose the well-known L2L method from machine learning. 
We found that for the case of reservoirs consisting of spiking neurons this two-tier process does in fact enhance subsequent reservoir learning performance substantially in terms of precision and speed of learning. 
We propose that similar advantages can be gained for other types of reservoirs, e.g. recurrent networks of artificial neurons or physical embodiments of reservoirs (see~\cite{tanaka2019recent} for a recent review) for which some of their parameters can be set to specific values. 
If one does not have a differentiable computer model for such physically implemented reservoir, one would have to use a gradient-free optimization method for the outer loop, such as simulated annealing or stochastic search, see \cite{BohnstinglETAL:19} for a first step in that direction.  

We have explored in section~\ref{sec:ltl-static} a variant of this method, where not even the weights to readout neurons need to be adapted for learning a specific tasks. 
Instead, the weights of recurrent connections within the reservoir can be optimized so that the reservoir can learn a task from a given family $\FF$ of tasks by maintaining learnt information for the current task in its working memory, i.e., in its network state. 
This state may include values of hidden variables such as current values of adaptive thresholds, as in the case of LSNNs \cite{bellec_etal_nips_2018}.
It turns out that L2L without any synaptic plasticity in the inner loop enables the reservoir to learn faster than the optimal learning method from machine learning for the same task: Backpropagation applied directly to the target network architecture which generated the nonlinear transformation, compare panels C and G of figure~\ref{fig:ltl2-results}.
We also have demonstrated in Fig~\ref{fig:ltl2-results}E (and in \cite{bellec_etal_nips_2018}) that the L2L method can be viewed as installing a prior in the reservoir. 
This observation raises the question what types of priors or rules can be installed in reservoirs with this approach. 
For neurorobotics applications it would be especially important to be able to install safety rules in a neural network controller that can not be overridden by subsequent learning. 
We believe that L2L methods could provide valuable tools for that.

Another open question is whether biologically more plausible and computationally more efficient approximations to BPTT, such as e-prop~\cite{bellec_solution_2019}, can be used instead of BPTT for optimizing a reservoir in the outer loop of L2L. In addition it was shown in~\cite{bellec_biologically_2019} that if one allows that the reservoir adapts weights of synaptic connections within a recurrent neural network via e-prop, even one-shot learning of new arm movements becomes feasible.

\subsubsection*{Acknowledgements}
This research was partially supported by the Human Brain Project, funded from the European Union's Horizon 2020 Framework Programme for Research and Innovation under the Specific Grant Agreement No. 720270 (Human Brain Project SGA1) and under the Specific Grant Agreement No. 785907 (Human Brain Project SGA2). 
Research leading to these results has in parts been carried out on the Human Brain Project PCP Pilot Systems at the J{\"u}lich Supercomputing Centre, which received co-funding from the European Union (Grant Agreement No. 604102).
We gratefully acknowledge Sandra Diaz, Alexander Peyser and Wouter Klijn from the Simulation Laboratory Neuroscience of the J{\"u}lich Supercomputing Centre for their support.

\bibliographystyle{apalike}
\bibliography{references}

\end{document}